\documentclass{article}

    \PassOptionsToPackage{numbers, compress}{natbib}

    \usepackage[preprint]{neurips_2020}

\usepackage{amsmath}
\usepackage{times}
\usepackage{url}
\usepackage{amssymb}
\usepackage{xcolor}
\usepackage{latexsym}
\usepackage{subfig}
\usepackage{varwidth}
\usepackage{graphicx}
\usepackage{cases}
\usepackage{placeins}
\usepackage{array, boldline, makecell, booktabs}
\usepackage{multirow}
\usepackage{multicol}
\usepackage{latexsym}
\usepackage{graphicx}
\usepackage{placeins}
\usepackage{booktabs}

\newcommand{\comment}[1]{}
\makeatletter
\def\thickhline{%
  \noalign{\ifnum0=`}\fi\hrule \@height \thickarrayrulewidth \futurelet
   \reserved@a\@xthickhline}
\def\@xthickhline{\ifx\reserved@a\thickhline
               \vskip\doublerulesep
               \vskip-\thickarrayrulewidth
             \fi
      \ifnum0=`{\fi}}
\makeatother
\usepackage{xcolor,colortbl}
\newlength{\thickarrayrulewidth}
\definecolor{Gray}{gray}{0.85}
\usepackage{subfig}
\usepackage{bm}

\usepackage[utf8]{inputenc} 
\usepackage[T1]{fontenc}    
\usepackage{hyperref}       
\usepackage{url}            
\usepackage{booktabs}       
\usepackage{amsfonts}       
\usepackage{nicefrac}       
\usepackage{microtype}      

\definecolor{blue_c}{RGB}{40, 116, 166}
\definecolor{orange_c}{RGB}{175, 96, 26 }

\newcolumntype{?}{!{\vrule width 3\arrayrulewidth}}

\title{STEPs-RL: Speech-Text Entanglement for Phonetically Sound Representation Learning}

\author{%
  Prakamya Mishra \\
  Independent Researcher\\
  \texttt{pkms.research@gmail.com} \\
}

\begin{document}

\maketitle

\begin{abstract}
  In this paper, we present a novel multi-modal deep neural network architecture that uses speech and text entanglement for learning phonetically sound spoken-word representations. STEPs-RL is trained in a supervised manner to predict the phonetic sequence of a target spoken-word using its contextual spoken word's speech and text, such that the model encodes its meaningful latent representations. Unlike existing work, we have used text along with speech for auditory representation learning to capture semantical and syntactical information along with the acoustic and temporal information. The latent representations produced by our model were not only able to predict the target phonetic sequences with an accuracy of 89.47\% but were also able to achieve competitive results to textual word representation models, Word2Vec \& FastText (trained on textual transcripts), when evaluated on four widely used word similarity benchmark datasets. In addition, investigation of the generated vector space also demonstrated the capability of the proposed model to capture the phonetic structure of the spoken-words. To the best of our knowledge, none of the existing works use speech and text entanglement for learning spoken-word representation, which makes this work first of its kind.
\end{abstract}

\section{Introduction}
Speaking and listening are the most common ways in which humans convey and understand each other in daily conversations. Nowadays, the speech interface has also been widely integrated into many applications/devices like Siri, Google Assistant, and Alexa \cite{herff2016automatic}. These applications use speech recognition-based approaches \cite{graves2013speech,jelinek1997statistical,bourlard2012connectionist} to understand the spoken user queries. Like speech, the text is also a widely used medium in which people converse. Recent advances in language modeling and representation learning using deep learning approaches \cite{mikolov2013distributed, bojanowski2017enriching, devlin-etal-2019-bert, peters-etal-2018-deep} have proven to be very promising in understanding the actual meanings of the textual data, by capturing semantical, syntactical, and contextual relationships between the textual words in their corresponding learned fixed-size vector representations. 

Such computational language modeling is difficult in the case of speech for spoken language understanding because unlike textual words, (1) spoken words can have different meanings of the same word when spoken in different tones/expressions \cite{glass1999challenges}, (2) it is difficult to identify sub-word units in speech because of the variable-length spacing and overlapping between the spoke-words \cite{vincent2013second}, and (3) use of stress/emphasis on few syllables of a multi-syllabic word can increase the variability of speech production \cite{polka2017segmenting}. Although the textual word representations capture the semantical, syntactical, and contextual properties, they fail to capture the tone/expression. Using only speech/audio data for training spoken-word representations results in semantically and syntactically poor representations. 

So in this paper, we propose a novel spoken-word representation learning approach called STEPs-RL that uses speech and text entanglement for learning phonetically sound spoken-word representations, which not only captures the acoustic and contextual features but also are semantically, syntactically, and phonetically sound. STEPs-RL is trained in a supervised manner such that the learned representations can capture the phonetic structure of the spoken-words along with their inter-word semantic, syntactic, and contextual relationships. We validated the proposed model by (1) evaluating semantical and syntactical relationships between the learned spoken-word representations on four widely used word similarity benchmark datasets, and comparing its performance with the textual word representations learned by Word2Vec \& FastTexT (obtained using transcriptions), and (2) investigating the phonetical soundness of the generated vector space.

\section{Related Work}
Earlier, speech processing was done using feature learning-based models like deep neural networks (DNN) \cite{8678825}. The DNN models were able to capture contextual and temporal information from the speech-based data after the introduction of sequential neural networks like RNNs \cite{miao2015eesen, hori2018end, rao2017exploring}, LSTMs \cite{li2015lstm, moriya2018lstm}, Bi-LSTMs \cite{zeyer2017comprehensive, graves2013hybrid}, and GRUs \cite{tang2017memory, ravanelli2018light}. 

Recent research by \citet{tera} has presented the use of a transformer-based self-supervised speech representation learning approach called TERA that uses multi-target auxiliary tasks. TERA is trained by generating acoustic frame reconstructions; \citet{Schneider2019} introduced wav2vec which is a CNN based model pre-trained in a unsupervised manner using contrastive loss to learn raw audio representations; \citet{khurana2020convolutional} explored the use of black-box variational inference for linguistic representation learning of speech using an unsupervised generative model; \citet{DBLP:journals/corr/abs-1807-03748} proposed Contrastive Predictive Coding (CPC) for extracting representations from high dimension data by predicting future in latent space, using autoregressive models; \citet{pmlr-v48-amodei16} presented an end-to-end deep learning model to recognize speech in two vastly different languages (English \& Mandarin Chinese); \citet{hsu2017learning} proposed a novel variational autoencoder based model that learns disentangled and interpretable latent representations of sequential data in an unsupervised manner; \citet{Ling2020} used BERT encoder for learning phonetically aware contextual speech representation vectors; \citet{8736337} proposed a Word2Vec type sequence-to-sequence autoencoder model for embedding variable-length audio segments. Other works on learning fixed-length spoken-word vector representations that use multi-task learning include \cite{10.1109/TASLP.2019.2938863}, \cite{8683639}, \cite{7404803}, \cite{Li2015ModelingSV} and \cite{Tan2016SpeakerawareTO}.

\section{Model}
\begin{figure}[t] 
    \centering
    \includegraphics[width=0.45\textwidth]{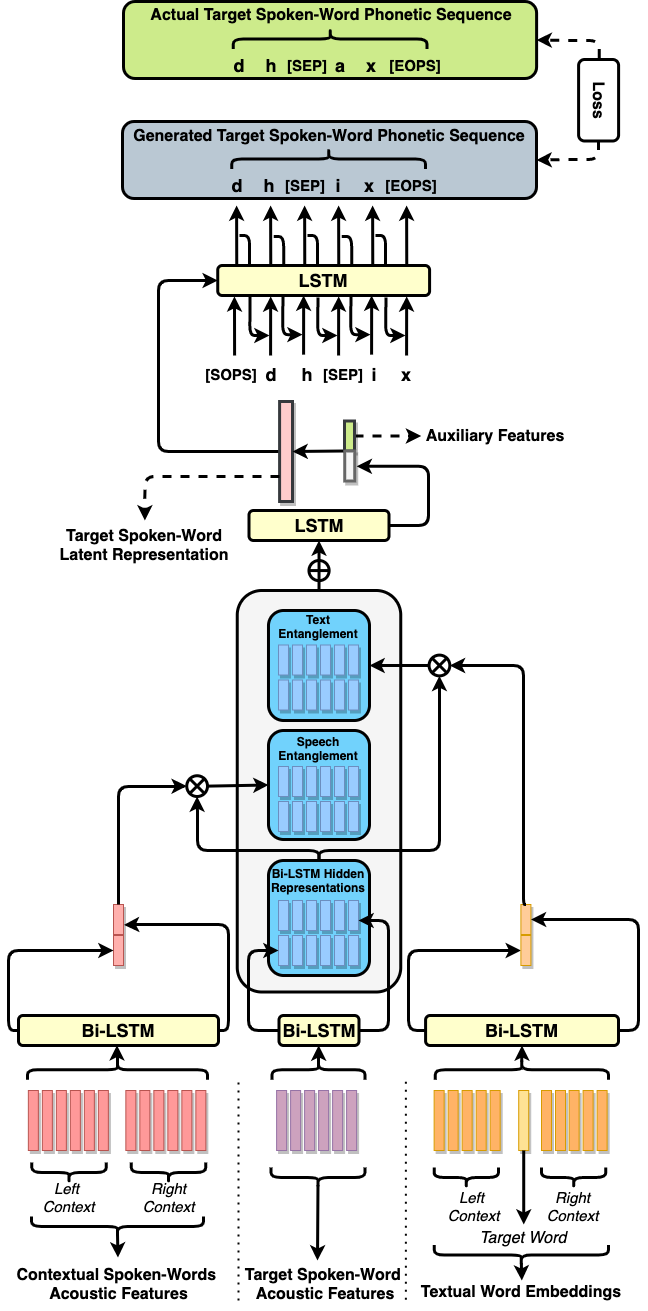}
    \caption{Illustration of the STEPs-RL model architecture.}
    \label{fig:main_model}
\end{figure}
In this paper, we propose STEPs-RL: Speech-Text Entanglement for Phonetically Sound Representation Learning. STEPs-RL is a novel spoken-word representation learning approach which entangles speech and text based contextual information for learning phonetically sound spoken-word representations. The model architecture is shown in Figure \ref{fig:main_model}. Given a target spoken-word represented by \(S^t\), its left and right contextual spoken-words represented by \(S_{ctx}^{l} = \{S^i\}_{t-1-m}^{t-1}\) \& \(S_{ctx}^{r} = \{S^i\}_{t+1}^{t+1+m}\) respectively (\(m\) represents the context window size), along with the textual word embeddings of the corresponding spoken-words represented by \(W_{ctx}^{l} = \{W^i\}_{t-1-m}^{t-1}\), \(W^t\) \& \(W_{ctx}^{r} = \{W^i\}_{t+1}^{t+1+m}\), the proposed model tries to learn a vector representation of the target spoken-word that not only captures the semantic-based, syntax-based and acoustic-based information but also captures the phonetic-based information.

Here, a single spoken-word \(S^i \in \mathbb{R}^{n\times d_{mfcc}}\) consists of a sequence of acoustic features represented by \(d_{mfcc}\)-dimensional Mel-frequency Cepstral Coefficients (MFCCs); \(W^i \in \mathbb{R}^{d_{w}}\) represents the \(d_{w}\)-dimensional pre-trained textual word embedding of the corresponding spoken-word. Each of the spoken-word is padded with silence, so that they all consists of a sequence of \(n\) acoustic features. 

Our approach uses Bidirectional-LSTM \cite{650093} for capturing the contextual information. Bidirectional-LSTM (also known as Bi-LSTM), uses two LSTM \cite{10.1162/neco.1997.9.8.1735} networks (\(\overrightarrow{LSTM}, \overleftarrow{LSTM}\)) to capture contextual information in opposite directions (forward and backward) of a sequence (\(t_1,..t_T\)). The final hidden representations corresponding to the sequence tokens is generated by concatenating (\(\oplus\)) the hidden representations (\(\overrightarrow{h_i},\overleftarrow{h_i}\)) generated by both the LSTM networks. So the final hidden representation of the \(i^{th}\) token can be represented as shown in equation \ref{equ2}.

\begin{equation} \label{equ1}
\overrightarrow{h_i} = \overrightarrow{LSTM}(t_i,\overrightarrow{h_{i-1}}), \quad  \overleftarrow{h_i} = \overleftarrow{LSTM}(t_i,\overleftarrow{h_{i + 1}})
\end{equation}
\begin{equation} \label{equ2}
h_i = \overrightarrow{h_i} \oplus \overleftarrow{h_i}
\end{equation}

STEPs-RL consist of three independent Bi-LSTM networks represented by \(BiLSTM_{C}\), \(BiLSTM_{T}\) and \(BiLSTM_{W}\) to capture contextual information respectively from (1) The acoustic features of the left and right contextual spoken-words represented by \(S_{ctx}^{l}\) \& \(S_{ctx}^{r}\), (2) The acoustic features of the target spoken-word represented by \(S^t\), and (3) The pre-trained textual word embeddings of the corresponding target spoken-word, left contextual spoken-words and right contextual spoken-words represented by \(W^t\), \(W_{ctx}^{l}\) \& \(W_{ctx}^{r}\) respectively.

\begin{equation} \label{equ3}
h^{C},\overrightarrow{o^{C}},\overleftarrow{o^{C}} = BiLSTM_{C}([S_{ctx}^{l},S_{ctx}^{r}])
\end{equation}
\begin{equation} \label{equ4}
h^{T},\overrightarrow{o^{T}},\overleftarrow{o^{T}} = BiLSTM_{T}([S^{t}])
\end{equation}
\begin{equation} \label{equ5}
h^{W},\overrightarrow{o^{W}},\overleftarrow{o^{W}}=BiLSTM_{W}([W_{ctx}^{l},W^t,W_{ctx}^{r}])
\end{equation}


As shown in equations \ref{equ3}, \ref{equ4} \& \ref{equ5}, all the three Bi-LSTM networks generate a final hidden state representation corresponding to each timestamp (\(h^{C}\), \(h^{T}\), \(h^{W}\)), a final output of the corresponding forward LSTM network (\(\overrightarrow{o^{C}}\), \(\overrightarrow{o^{T}}\), \(\overrightarrow{o^{W}}\)), and a final output of the corresponding backward LSTM network (\(\overleftarrow{o^{C}}\), \(\overleftarrow{o^{T}}\), \(\overleftarrow{o^{W}}\)). The final forward and backward outputs of \(BiLSTM_{C}\) \& \(BiLSTM_{W}\) are concatenated to generate \(f^C\) \& \(f^W\) respectively, which will later act as context vectors during the entanglement of speech and text. 
\begin{equation} \label{equ6}
f^C = \overrightarrow{o^{C}} \oplus \overleftarrow{o^{C}}, \quad f^W = \overrightarrow{o^{W}} \oplus \overleftarrow{o^{W}}
\end{equation}

\begin{figure}[t]%
    \centering
    \subfloat[]{{\includegraphics[width=0.5\textwidth]{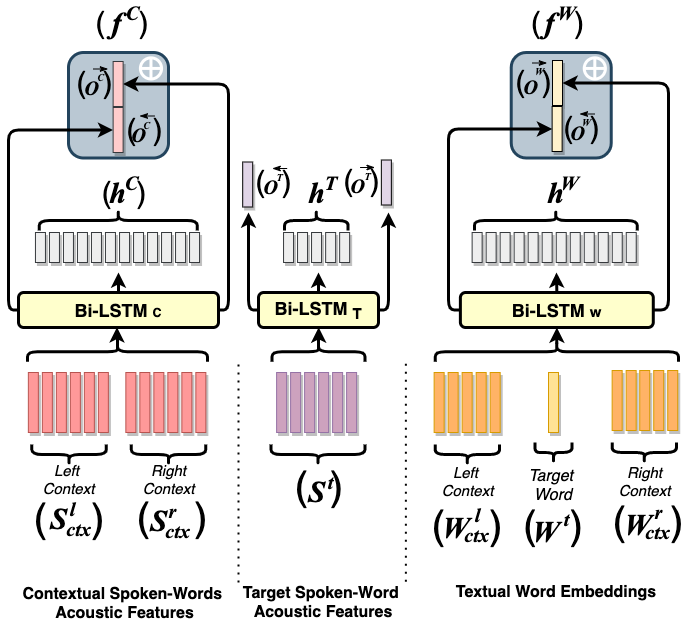} }\label{fig:sec1}}%
    \qquad
    \subfloat[]{{\includegraphics[width=0.4\textwidth]{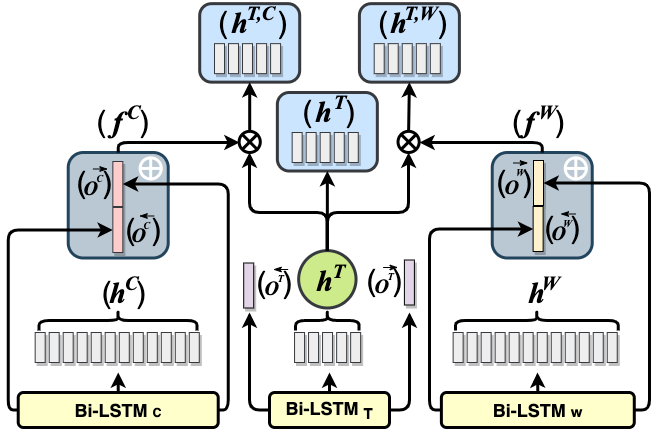} }\label{fig:sec2}}%
    \caption{(a) STEPs-RL Phase 1: Each of the individual Bi-LSTM captures contextual information. (b) STEPs-RL Phase 2: Speech \& Text entanglement with target spoken word.}
\end{figure}

For intuition (as shown in Figure \ref{fig:sec1}), \(f^C\) represents the final contextual representation of the spoken-words present in context of the target spoken-word, and \(f^W\) represents the final semantical and syntactical contextual representation of all the corresponding textual words. In other words, \(f^C\) captures the acoustic/speech-based contextual information whereas \(f^W\) captures the text-based contextual information. Both \(f^C\) \& \(f^W\), are then used to entangle speech and text-based contextual information with the target spoken-word by generating new speech and text entangled bidirectional hidden state representations (\(h^{T,C}\) \& \(h^{T,W}\)) of the target spoken-word using the hidden representations generated by \(BiLSTM_{T}\), as shown in equations \ref{equ7}, \ref{equ8}, \ref{equ9} \& \ref{equ10}.

\begin{equation} \label{equ7}
h^{T,C} = [h_1^{T,C}, h_2^{T,C},...,h_n^{T,C}] = h^{T} \otimes f^C
\end{equation}
\begin{equation} \label{equ8}
h_i^{T,C} = \alpha_i^{T,C} \times h_i^{T}
\end{equation}
\begin{equation} \label{equ9}
h^{T,W} = [h_1^{T,W}, h_2^{T,W},...,h_n^{T,W}] = h^{T} \otimes f^W
\end{equation}
\begin{equation} \label{equ10}
h_i^{T,W} = \alpha_i^{T,W} \times h_i^{T}
\end{equation}

In the above equations, (\(\otimes\)) represents an element wise attention function; \(h^{T,C}\) \& \(h^{T,W}\) represents the newly generated speech-entangled and text-entangled hidden representations respectively; \(\alpha_i^{T,C}\) \& \(\alpha_i^{T,W}\) represents the speech-entangled and text-entangled attention scores respectively, corresponding to the \(i^{th}\) timestamp of the hidden representations generated by \(BiLSTM_{T}\). The attention scores \(\alpha_i^{T,C}\) \& \(\alpha_i^{T,W}\) are generated by taking the dot product (\(\bullet\)) of each of the timestamps of \(h^T\) with the context vectors \(f^C\) \& \(f^W\) respectively, as shown in equation \ref{equ11}. Same is illustrated in Figure \ref{fig:sec2}.

\begin{equation} \label{equ11}
\alpha_i^{T,C} = h_i^{T} \bullet f^C, \quad \alpha_i^{T,W} = h_i^{T} \bullet f^W
\end{equation}

\begin{figure}[t] 
    \centering
    \includegraphics[width=0.35\textwidth]{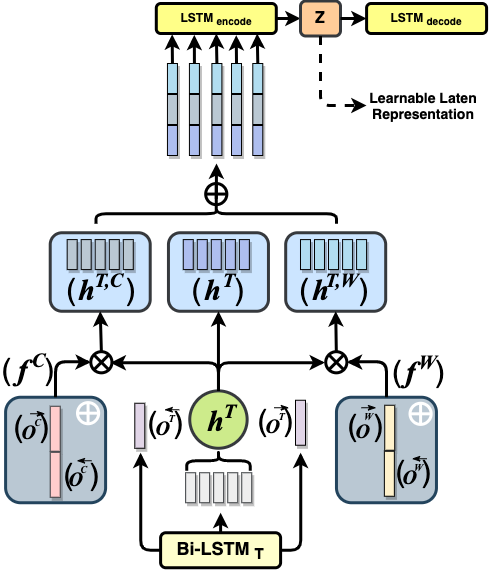}
    \caption{STEPs-RL Phase 3: Latent representation learning}
    \label{fig:sec3}
\end{figure}
Next, the proposed model uses the newly generated speech-entangled and text-entangled hidden representations \(h^{T,C}\) \& \(h^{T,W}\), along with the original bidirectional hidden state representations \(h^T\) of the target spoken-word (generated from \(BiLSTM_{T}\)), to generate a latent vector representation \(z\) of the target spoken-word by stacking (illustrated in Figure \ref{fig:sec3}) all these three hidden representations on top of each other and passing it through a simple encoder LSTM network \(\overrightarrow{LSTM_{encode}}\).

\begin{equation} \label{equ12}
z = \overrightarrow{LSTM_{encode}}([h^{T,C} \oplus h^{T,W} \oplus h^T])
\end{equation}
\begin{equation} \label{equ13}
z_{new} = zW_1 + z_{aux}W_2 + B
\end{equation}

In equation \ref{equ12}, \(z\) represents a fixed size latent vector which is the output of the encoder LSTM network. To add more information about the speaker, the proposed model linearly combines the latent vector with an auxiliary vector \(z_{aux}\) to generate a new latent representation \(z_{new}\) of the target spoken-word. This new latent representation \(z_{new} \in \mathbb{R}^{d_{e}}\) is a \(d_{e}\)-dimensional vector representation that the proposed model tries to learn. In equation \ref{equ13}, \(W_1 \in \mathbb{R}^{d\times d_{e}}\) and \(W_2  \in \mathbb{R}^{d_a\times d_{e}}\) represents the combination weights and \(B\) represents the bias. These weights and biases are learnable in nature. The auxiliary vector \(z_{aux} \in \mathbb{R}^{d_a}\) is a one-hot vector of size \(d_a\)that consists of information related to the speaker's gender/dialect or both. Such an auxiliary vector was introduced because usually, the pronunciation of different words usually depends on the speaker's gender and dialect and hence can help learn phonetically sound spoken-word representations. 

Next, the proposed model uses a decoder LSTM network \(\overrightarrow{LSTM_{decode}}\) to predict the sequence of phonetic symbols \(Y\)=(\([y_1,...,y_k]\)) of the corresponding target spoken-word using the above generated latent representation of the target spoken-word \(z_{new}\), as shown in equation \ref{equ14} \& \ref{equ15}. 

\begin{equation} \label{equ14}
P_{\theta}(y_i|Y_{<i}, z_{new}) = \Upsilon(h_i^d, y_{i-1})
\end{equation}
\begin{equation} \label{equ15}
h_i^d = \Psi(h_{i-1}^d, y_{i-1})
\end{equation}

Here, \(\Psi\) represents a function that generates the hidden vectors \(h_i^d\) (hidden state representations of the decoder network), and \(\Upsilon\) represents a function that computes the generative probability of the one-hot vector \(y_i\) (target phonemic symbol). The hidden vector \(h_i^d\) is \(z_{new}\), and \(y_i\) is the one-hot vector of "[SOP]" when \(i\)=\(0\). Here "[SOP]" represent the start of phoneme token. The proposed model uses cross-entropy as its training loss function as shown in equation \ref{equ16}, where cross-entropy loss \(L\) is computed using the actual target spoken-word phonetic sequence (\(Y\)=(\([y_1,...,y_k]\))) and the predicted target spoken-word phonetic sequence (\(\hat{Y}\)=(\([\hat{y_1},...,\hat{y_k}]\))).

\begin{equation} \label{equ16}
L(Y,\hat{Y}) = \sum_{i}^{k}y_i\log \frac{1}{\hat{y_i}}
\end{equation}

\section{Experimental Setup}

\subsection{Dataset}

For our experiments, we used the DARPA TIMIT acoustic-phonetic Continuous Speech Corpus \cite{garofolo1993darpa}. This corpus contains 16kHz audio recordings of 630 speakers of 8 major American English dialects of which approximately 70\% were male and 30\% were female, as shown in Table \ref{tab:d}. The corpus consists of 6300 (5.4 hours) phonetically rich utterances by different speakers (10 by each speaker) along with their corresponding time-aligned orthographic, phonetic, and word transcriptions. All the recordings were segmented according to the spoken-word boundaries using the transcriptions and were paired with their left and right context spoken-words and their corresponding textual words along with the phonetic sequence of the target spoken-word. All the spoken-word utterances were represented by their MFCC representations and the textual words were represented by their pre-trained textual word embeddings, where the MFCC representations and the textual word embeddings were of the same size (\(d_{mfcc}\)=\(d_{w}\)). One-hot encoded dialect (8-dimensional) and gender (2-dimensional) vectors were used as auxiliary information vectors.

\subsection{Training Details}

We used the standard train (462 speakers and 4956 utterances) and test (168 speakers and 1344 utterances) set of the TIMIT speech corpus for training and testing the proposed model. Due to computational resource limitations, a context window size of 3 was used. In all the experiments the MFCC representations and the textual word embeddings were of the same size (\(d_{mfcc}\)=\(d_{w} \in\) \{50,100,300\}). For the textual word embeddings, the proposed model used two different widely used pre-trained word embeddings i.e, (1) Word2Vec \cite{mikolov2013distributed}, which are word-based embeddings, and (2) FastText \cite{bojanowski2017enriching}, which are character-based embeddings. For all the experiments, the proposed model was trained for 20 epochs using a mini-batch size of 100. The initial learning rate was set to \(0.01\) and Adam optimizer was used for optimization. The Bi-LSTM and LSTM nodes were regularised using an L2 regularizer with a penalty of \(0.01\). Early stopping was used to avoid over-fitting. The size of the target spoken-word latent representation \(z_{new}\) was set to 50-, 100- \& 300 for comparison. All the spoken-words were represented by a sequence of 50 phonetic symbols using the original unique 27 phonetic symbols present in the corpus along with our four newly introduced symbols ("\textbf{[SOPS]}" for the start of each phonetic sequence, "\textbf{[SEP]}" for separation/space between phonetic symbols, "\textbf{[PAD]}" for padding and "\textbf{[EOPS]}" for the end of each phonetic sequence). 
\begin{table}
\caption{Gender and dialect distribution of the speakers in TIMIT speech corpus.}
\label{tab:d}
\centering
\resizebox{0.6\textwidth}{!}{%
\begin{tabular}{ccccccccc}
\hline
 & \multicolumn{8}{c}{Dialect} \\ \cline{2-9} 
Gender & 1 & 2 & 3 & 4 & 5 & 6 & 7 & 8 \\ \hline \hline
\multicolumn{1}{c}{Male} & 63\% & 70\% & 67\% & 69\% & 63\% & 65\% & 74\% & 67\% \\
\multicolumn{1}{c}{Female} & 27\% & 30\% & 27\% & 31\% & 37\% & 35\% & 26\% & 33\% \\ \hline
\multicolumn{1}{c}{Total} & \textbf{8\%} & \textbf{16\%} & \textbf{16\%} & \textbf{16\%} & \textbf{16\%} & \textbf{7\%} & \textbf{16\%} & \textbf{5\%} \\ \hline
\end{tabular}%
}
\end{table}
\section{Results}

\begin{table}
\caption{Phonetic sequence prediction results on the TIMIT speech corpus. We present here the comparison of testing set accuracy (\%) of the STEPs-RL model using different sets of auxiliary information (gender (\textbf{G}), dialect (\textbf{D})) with the base STEPs-RL model using no auxiliary information. The comparison is done for different textual word embedding sizes \(\bm{d_{w}}\)=\{50,100,300\}, different spoken-word latent representation sizes \(\bm{d_{e}}\)=\{50,100,300\} and different word embeddings like Word2Vec (\textit{\textbf{w}}) and FastText (\textit{\textbf{f}}). The best performance in each configuration is marked in \textbf{bold}, row of the best performing model is highlighted in \textbf{\colorbox{Gray}{grey}}, the overall best performance is further marked in \textbf{\textcolor{red}{red}} and its configuration is marked in \textbf{\textcolor{blue}{blue}}.}
\label{tab:r1}
\centering
\resizebox{\textwidth}{!}{%
\begin{tabular}{cccccccccccccccccccc}
\hline
\multirow{2}{*}{\textbf{Spoken-Word Latent Rep. Size (\(\bm{d_{e}}\))}} & \multirow{2}{*}{$\rightarrow$} & \multicolumn{6}{c}{\multirow{2}{*}{\large{\textbf{\(\bm{d_{e}}\) = 50}}}} & \multicolumn{6}{c}{\multirow{2}{*}{\large{\textbf{\(\bm{d_{e}}\) = 100}}}} & \multicolumn{6}{c}{\multirow{2}{*}{\textbf{\textcolor{blue}{\large{\textbf{\(\bm{d_{e}}\) = 300}}}}}} \\  
 & & \multicolumn{6}{c}{} & \multicolumn{6}{c}{} & \multicolumn{6}{c}{} \\\cmidrule(lr){3-8} \cmidrule(lr){9-14} \cmidrule(lr){15-20}
\multirow{2}{*}{\textbf{Textual Word Embedding Size (\(\bm{d_{w}}\))}} & \multirow{2}{*}{$\rightarrow$} & \multicolumn{2}{c}{\multirow{2}{*}{\large{\textbf{\(\bm{d_{w}}\) = 50}}}} & \multicolumn{2}{c}{\multirow{2}{*}{\large{\textbf{\(\bm{d_{w}}\) = 100}}}} & \multicolumn{2}{c}{\multirow{2}{*}{\large{\textbf{\(\bm{d_{w}}\) = 300}}}} &
\multicolumn{2}{c}{\multirow{2}{*}{\large{\textbf{\(\bm{d_{w}}\) = 50}}}} & \multicolumn{2}{c}{\multirow{2}{*}{\large{\textbf{\(\bm{d_{w}}\) = 100}}}} & \multicolumn{2}{c}{\multirow{2}{*}{\large{\textbf{\(\bm{d_{w}}\) = 300}}}} &
\multicolumn{2}{c}{\multirow{2}{*}{\textbf{\textcolor{blue}{\large{\textbf{\(\bm{d_{w}}\) = 50}}}}}} & \multicolumn{2}{c}{\multirow{2}{*}{\large{\textbf{\(\bm{d_{w}}\) = 100}}}} & \multicolumn{2}{c}{\multirow{2}{*}{\large{\textbf{\(\bm{d_{w}}\) = 300}}}} \\
 &  & \multicolumn{2}{c}{} & \multicolumn{2}{c}{} & \multicolumn{2}{c}{} & \multicolumn{2}{c}{} & \multicolumn{2}{c}{} & \multicolumn{2}{c}{} & \multicolumn{2}{c}{} & \multicolumn{2}{c}{} & \multicolumn{2}{c}{} \\ \cmidrule(lr){3-4} \cmidrule(lr){5-6} \cmidrule(lr){7-8} \cmidrule(lr){9-10} \cmidrule(lr){11-12} \cmidrule(lr){13-14} \cmidrule(lr){15-16} \cmidrule(lr){17-18} \cmidrule(lr){19-20} 
\multirow{2}{*}{\textbf{Textual Word Embeddings Used}} & \multirow{2}{*}{$\rightarrow$} & \multicolumn{1}{c}{\multirow{2}{*}{\large{\textit{\textbf{w}}}}} & \multicolumn{1}{c}{\multirow{2}{*}{\large{\textit{\textbf{f}}}}} & \multicolumn{1}{c}{\multirow{2}{*}{\large{\textit{\textbf{w}}}}} & \multicolumn{1}{c}{\multirow{2}{*}{\large{\textit{\textbf{f}}}}} & \multicolumn{1}{c}{\multirow{2}{*}{\large{\textit{\textbf{w}}}}} & 
\multicolumn{1}{c}{\multirow{2}{*}{\large{\textit{\textbf{f}}}}} & \multicolumn{1}{c}{\multirow{2}{*}{\large{\textit{\textbf{w}}}}} & \multicolumn{1}{c}{\multirow{2}{*}{\large{\textit{\textbf{f}}}}} & \multicolumn{1}{c}{\multirow{2}{*}{\large{\textit{\textbf{w}}}}} & \multicolumn{1}{c}{\multirow{2}{*}{\large{\textit{\textbf{f}}}}} & \multicolumn{1}{c}{\multirow{2}{*}{\large{\textit{\textbf{w}}}}} &
\multicolumn{1}{c}{\multirow{2}{*}{\large{\textit{\textbf{f}}}}} & \multicolumn{1}{c}{\multirow{2}{*}{\textbf{\textcolor{blue}{\large{\textit{\textbf{w}}}}}}} & \multicolumn{1}{c}{\multirow{2}{*}{\large{\textit{\textbf{f}}}}} & \multicolumn{1}{c}{\multirow{2}{*}{\large{\textit{\textbf{w}}}}} & \multicolumn{1}{c}{\multirow{2}{*}{\large{\textit{\textbf{f}}}}} & \multicolumn{1}{c}{\multirow{2}{*}{\large{\textit{\textbf{w}}}}} &
\multicolumn{1}{c}{\multirow{2}{*}{\large{\textit{\textbf{f}}}}}\\
  &  & \multicolumn{1}{c}{} & \multicolumn{1}{c}{} & \multicolumn{1}{c}{} & \multicolumn{1}{c}{} & \multicolumn{1}{c}{} & \multicolumn{1}{c}{} & \multicolumn{1}{c}{} & \multicolumn{1}{c}{} & \multicolumn{1}{c}{} & \multicolumn{1}{c}{} & \multicolumn{1}{c}{} &\multicolumn{1}{c}{} & \multicolumn{1}{c}{} & \multicolumn{1}{c}{} & \multicolumn{1}{c}{} & \multicolumn{1}{c}{} & \multicolumn{1}{c}{} &\multicolumn{1}{c}{}
\\ 
\hline \hline
\textbf{STEPs-RL + No auxiliary information} & & 71.67 & 73.24 & 75.22 & 78.41 & 84.76 & 84.98 & 73.44 & 76.72 & 78.36 & 81.23 & 86.90 & 86.92 & 75.22 & 79.75 & 80.59 & 83.73 & 87.31 & 86.90 \\
\textbf{STEPs-RL + D} & & 83.87 & 84.01 & 86.11 & 86.36 & 85.89 & 87.89 & 86.83 & 86.85 & 86.75 & 87.54 & 87.23 & 88.00 & 81.50 & 82.05 & 85.90 & 86.36 & 87.40 & 87.99 \\
\textbf{STEPs-RL + G} & & 87.93 & 86.97 & 87.44 & 85.60 & 87.16 & 88.28 & 87.15 & 87.12 & 87.98 & \textbf{88.20} & \textbf{88.92} & \textbf{88.54} & 87.16 & 87.30 & 88.91 & 88.45 & 88.10 & 88.59 \\
\rowcolor{Gray}\textbf{STEPs-RL + D + G} & & \textbf{88.91} & \textbf{88.04} & \textbf{88.10} & \textbf{88.28} & \textbf{88.94} & \textbf{88.78} & \textbf{88.27} & \textbf{88.90} & \textbf{89.14} & 88.08 & 88.59 & 89.18 &  \textbf{\textcolor{red}{89.47}} & \textbf{89.21} & \textbf{89.35} & \textbf{89.38} & \textbf{88.63} & \textbf{89.41} \\
\hline
\end{tabular}%
}
\end{table}
For evaluation, we first tested the proposed model on the phonetic sequence prediction task with different spoken-word latent representation \& textual word embedding sizes, and also tested the performance of the model using different types of textual word embeddings (Word2Vec \& FastText). We compared the phonetic sequence prediction accuracy (\%) of the base STEPs-RL model (w/o any auxiliary information) with its variants that use different sets of auxiliary information like gender/dialect or both. The results are shown in Table \ref{tab:r1}. It was observed that increasing the spoken-word representation size resulted in better performance but was not so evident in the case of textual word embedding size. It was also observed that in general using Word2Vec textual word embeddings achieved better results compared to using FastText textual word embeddings. The addition of auxiliary information like dialect and gender showed clear improvements in accuracy when compared to the base STEPs-RL model, validating the use of this type of auxiliary information for spoken-word representation learning. It was also found that STEPs-RL was able to perform best when it used both dialect (\textbf{D}) and gender (\textbf{G}) together in its auxiliary vector (\textbf{STEPs-RL+D+G}). So for further evaluations, we will only consider the target spoken-word representations generated from the STEPs-RL+D+G model using the configurations marked blue in Table \ref{tab:r1}. Table \ref{tab:phonetics} illustrates examples of four different spoke-words along with their actual corresponding phonetic sequences and the phonetic sequences predicted by the STEPs-RL+D+G model. These examples demonstrate the ability of the STEPs-RL+D+G model to encode phonetic-based information in their corresponding latent representations.

\begin{table}%
    \caption{(a) Examples of the phonetic sequences generated by STEPs-RL+D+G model. (b) Performance of STEPs-RL+D+G compared to Word2Vec \& FastText on four benchmark word similarity datasets.}%
    \centering
    \subfloat[]{
    {\resizebox{0.42\textwidth}{!}{%
    \begin{tabular}{c|c|c}
    \hline
    \textbf{Ground Truth} & \textbf{Ground Truth} & \textbf{Generated} \\
    \textbf{Word} & \textbf{Phonetic Sequence} & \textbf{Phonetic Sequence} \\ \hline \hline
    that & dh ae tcl & dh ax tc l \\
    shelter & s sh eh l tcl t axr & s tsh ah el t er \\
    near & n ih axr & n ehh axr \\
    heck & hv eh kcl k & hv ah ncl k \\ \hline
    \end{tabular}%
    }}\label{tab:phonetics}
    }%
    \qquad
    \subfloat[]{
    {\resizebox{0.5\textwidth}{!}{%
    \begin{tabular}{c|c|c|c|c}
    \hline
    \textbf{Dataset} & \textbf{\# Word Pairs} & \textbf{STEPs + D + G ($\rho$)} & \textbf{Word2Vec ($\rho$)} & \textbf{FastText ($\rho$)} \\ \hline \hline
    \textbf{SimLex} &  136 & 0.2552 & 0.2792  & 0.3375 \\
    \textbf{MTurk} &  55 &  0.5312 & 0.6536 & 0.6771 \\
    \textbf{WS} & 17 &  0.2726 &  0.3248 & 0.3082 \\
    \textbf{Verb} &  26 &  0.3260 &   0.3508 &  0.3676 \\\hline
    \end{tabular}%
    }}\label{tab:sim-table}
    }%
\end{table}



To further evaluate the latent representations generated from STEPs-RL+D+G, we use intrinsic methods to test the semantic or syntactic relationships between these generated latent representations of the spoken-words present in the corpus. To do so, we use 4 benchmark word similarity datasets and compare the performance of the spoken-word representations generated from STEPs-RL+D+G with the representations generated by text-based language models (Word2Vec \& FastText) trained on the textual transcripts. The word similarity datasets include SimeLex-999 \cite{hill2015simlex}, MTurk-771 \cite{halawi2012large}, WS-353 \cite{yang2006verb} and Verb-143 \cite{baker2014unsupervised}. These datasets contain pairs of English words and their corresponding human-annotated word similarity ratings. The word similarities between the spoken-words (in case of STEPs-RL+D+G) and the textual-words (in case of Word2Vec and FastText) were obtained by measuring the cosine similarities between their corresponding representation vectors. 

Table \ref{tab:sim-table} reports Spearman’s rank correlation coefficient $\rho$ between the human rankings and the ones generated by STEPs-RL+D+G, Word2Vec, and Fastext. Since there were many words present in these datasets which were not present in the TIMIT speech corpus, only those word pairs were considered in which both the word were present in the TIMIT speech corpus. Table \ref{tab:sim-table} shows that the performance of the spoken-word representations generated from STEPs-RL+D+G was comparable to the performance of textual word representations generated from Word2Vec and FastText. This demonstrates that our proposed model was also able to capture semantic-based and syntax-based information, although the scores were slightly less compared to Word2Vec and FastText. 

We believe that the primary reason for this difference is the disparity in the way different speakers speak. The same word can be spoken in different ways and can have different meanings based on the tone and expression which may in return lead to an entirely different representation for the same word. In addition to that, these word similarity datasets are for the textual words, which do not take into account the tone and the expression aspect. Also, to the best of our knowledge, no other such word similarity dataset exists for the spoken-words. So keeping in mind these issues, the performance of the proposed model validates its ability to capture semantical and syntactical information in the representations it generates.

\begin{figure}[t]%
    \centering
    \subfloat[]{{\includegraphics[width=0.465\textwidth]{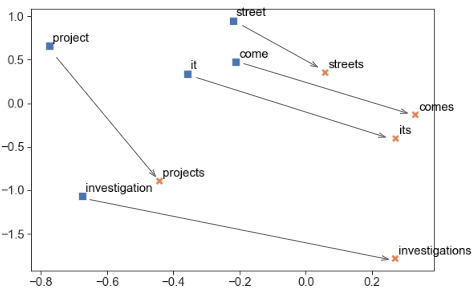} }\label{a}}%
    \qquad
    \subfloat[]{{\includegraphics[width=0.465\textwidth]{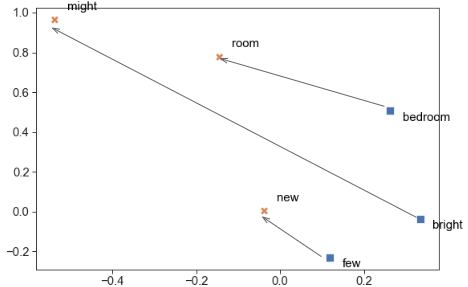}
    }\label{b}}%
    \caption{Difference vectors corresponding to (a) Set 1: Word pairs differ in last few phonemes (b) Set 2: Word pairs differ in first few phonemes}%
    \label{fig:example}%
\end{figure}
Next, we try to investigate the phonetical soundness of the vector space generated by the proposed model. A vector space can be said to be phonetically sound if the spoken-word representations of the words having similar pronunciations are present close to each other in the vector space. For this investigation we use 2 sets of randomly chosen word pairs:
\begin{itemize}
    \item \textbf{Set 1}: (street,streets), (come,comes), (it,its), (project,projects), (investigation,investigations)
    \item \textbf{Set 2}: (few,new), (bright,night), (bedroom,room)
\end{itemize}
Here, in Set 1 the word pairs differ in the last few phonemes and in Set 2 the word pairs differ in the first few phonemes. To illustrate the relationship between these word pairs, first, the difference vectors were computed between the average spoken-word vector representation of the words present in the above-mentioned word pairs, and then these high dimensional difference vectors were reduced to 2-dimensional vectors using PCA \citep{hotelling1933analysis}, to interpret these vectors.  The difference vectors corresponding to Set 1 \& Set 2 are shown in Figure \ref{fig:example}. It can be observed in the figures that the difference vectors are similar in directions and magnitude. In both the figures, phonetic replacements lead to similar transformations, for example (come \(\to\) comes) is similar to (it \(\to\) its) in Figure \ref{a}, and (few \(\to\) new) is similar to (bright \(\to\) night) in Figure \ref{b}. These transformations are not perfectly similar because we are taking an average of the same word spoken by different speakers having different accents and pronunciations, but despite this, the transformations are still very close to each other.

All these experiments demonstrate the quality of spoken-word vector representations generated by the proposed model using speech and text entanglement which not only are semantically and syntactically adequate but are also phonetically sound.

\section{Conclusion}
In this paper, we introduced STEPs-RL for learning phonetically sound spoken-word representations using speech and text entanglement. Our approach achieved an accuracy of 89.47\% in predicting phonetic sequences when both gender and dialect of the speaker are used in the auxiliary information. We also compared its performance using different configurations and observed that the performance of the proposed model improved by (1) increasing the spoken word latent representation size, and (2) the addition of auxiliary information like gender and dialect. We were not only able to validate the capability of the learned representations to capture the semantical and syntactical relationships between the spoken-words but were also able to illustrate soundness in the phonetic structure of the generated vector space. For future work, we plan to (1) extend the model to use attention mechanisms, (2) improve performance by using transformer-based architecture, and (3) experimenting on larger datasets, and (4) using features other than MFCCs.

\bibliography{anthology, neurips_2020}
\bibliographystyle{acl_natbib}

\end{document}